\documentclass[conference, letterpaper]{IEEEtran}
\usepackage{amsmath,amsfonts,amssymb,amsthm,epsfig,epstopdf,url,array}
\usepackage[table]{xcolor}
\usepackage{algorithmic}
\usepackage{balance}
\usepackage[ruled]{algorithm2e}
\usepackage{stmaryrd}
\usepackage{centernot}
\usepackage{tikz}
\usepackage{booktabs}
\usepackage{pdfpages}
\usepackage{graphicx}
\usepackage[font=small,skip=8pt]{caption}
\usepackage{tabularx}
\usepackage{comment}
\usepackage{multirow}
\usepackage{pdflscape}
\usepackage{rotating}
\usepackage{float}
\usetikzlibrary{plotmarks,shapes,arrows,chains,hobby,backgrounds,calc,trees}
\usepackage[top=1in, bottom=1in, left=1in, right=1in]{geometry}
\usepackage{makecell}
\usepackage{wrapfig,lipsum,booktabs}
\usepackage{ifpdf}
\usepackage{cite}

\usepackage{algorithmic}
\usepackage{array}
\usepackage{eqparbox}
\usepackage[tight,footnotesize]{subfigure}
\usepackage[font=footnotesize]{subfig}
\usepackage{stfloats}
\usepackage{url}

\begin{document}
\title{GOTM: a Goal-Oriented Framework for \\Capturing Uncertainty of Medical Treatments}

\author{\IEEEauthorblockN{Davoud Mougouei, David M. W. Powers}
	\IEEEauthorblockA{College of Science and Engineering, 
		Flinders University\\
		Adelaide, Australia\\
		\{davoud.mougouei,david.powers\}@flinders.edu.au}
}

\maketitle

\begin{abstract}
It has been widely recognized that uncertainty is an inevitable aspect of diagnosis and treatment of medical disorders. Such uncertainties hence, need to be considered in computerized medical models. The existing medical modeling techniques however, have mainly focused on capturing uncertainty associated with diagnosis of medical disorders while ignoring uncertainty of treatments. To tackle this issue, we have proposed using a fuzzy-based modeling and description technique for capturing uncertainties in treatment plans. We have further contributed a formal framework which allows for goal-oriented modeling and analysis of medical treatments.   
\end{abstract}

\IEEEpeerreviewmaketitle
\section{introduction}

It has been widely recognized in the literature that associations between symptoms and medical disorders are uncertain~\cite{griffiths2005nature,sonnenberg1993markov,adlassnig_fuzzy_1980,adlassnig_cadiag_1985}. In other words, it is not known for sure that a set of medical symptoms would indicate a certain medical disorder~\cite{adlassnig_fuzzy_1980}. In a similar spirit, associations between elements of a treatment plan (namely treatment goals and medical interventions)~\cite{rosen1991treatment} are also uncertain~\cite{politi2007communicating,pan_model_2015,ying_fuzzy_2006, belal_intelligent_2005}. In other words, employing medical interventions designed for satisfaction of a certain treatment goal may not fully satisfy that goal~\cite{politi_communicating_2007,mishel1988uncertainty,germino_uncertainty_1998,van2002inclusion}. Hence, efficient medical models must capture uncertainties in treatment plans (guidelines). 

Fuzzy logic, probability theory, and possibility theory have long been used in engineering~\cite {ghanooni2020robust} and  medical~\cite{owens2001medical,ciabattoni_formal_2013,adlassnig_cadiag_1985} expert systems for capturing uncertainty of medical diagnosis. The existing techniques for treatment modeling~\cite{ten_computer_2008} however, have mainly ignored considering uncertainties in medical plans. Ying et. al~\cite{ying_fuzzy_2006} underlined uncertainty of HIV (Human Immunodeficiency Virus) treatment plans by proposing a fuzzy discrete event system for capturing uncertainty in finding an optimal regiment for treatment of HIV infection. Although the work underlines captures uncertainty associated with selection of medical interventions~\cite{meltzer2001addressing} it fails to consider uncertainties of medical interventions themselves.

In this paper, we have proposed using fuzzy logic for capturing uncertainty in treatment modeling of medical disorders. We have further contributed a goal-oriented framework referred to as \textit{GOTM (Goal-Oriented Treatment Modeling)} which allows for goal-oriented modeling and description of medical treatments. Goal-oriented approach toward treatment modeling in \textit{GOTM} provides structured arguments for choosing medical interventions~\cite{ten_computer_2008} that satisfy certain treatment goals~\cite{ten_computer_2008}. Fuzzy algebraic structure of treatment models in \textit{GOTM} allows for capturing uncertainty of treatment goals and medical interventions in medical plans. 

The remainder of this paper is organized as follows. Section~\ref{sec_gotm} presents \textit{GOTM} framework and its application on a hypothetical treatment plan, and Section~\ref{sec_conclusion} concludes our work with a summary of major contributions and future work.

\section{The \textit{GOTM} Framework}
\label{sec_gotm}

\begin{figure}[!b]
	\centering
	\includegraphics[scale=0.6]{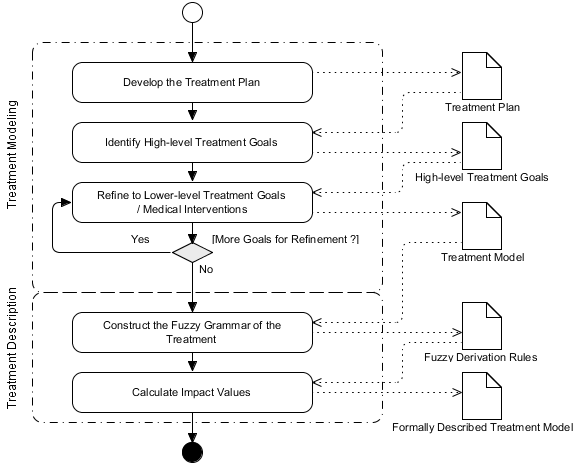}
	\caption{The \textit{GOTM} Framework}
	\label{fig_framework}
\end{figure}

Based on our earlier works in software requirement engineering~\cite{mougouei2012goal,mougouei2013fuzzyBased} we have contributed \textit{GOTM} in order to consider uncertainty in development of medical treatment plans (guidelines). As depicted in Figure~\ref{fig_framework}, \textit{GOTM} consists of two main components: 1) the treatment modeling process, and 2) the treatment description process. 

The process starts with developing a treatment model~\cite{weiss1978model}, that will be formally described to allow for an automated analysis of treatment plans. 

\subsection{Modeling Treatment Guidelines}
\label{sec_gotm_modeling}
As explained earlier, treatment plans (guidelines) are technical documents which capture treatment goals and medical interventions required for satisfaction of those goals~\cite{ten_computer_2008}. Various techniques~\cite{fox1998disseminating,shahar1998asgaard,ohno1998guideline,clayton1989issues} could be used for development of treatment plans. Modeling and description processes in \textit{GOTM} nevertheless, are independent of the  approach used for development of treatment plans. 

\begin{figure*}[!t]
\centering
\includegraphics[scale=0.5]{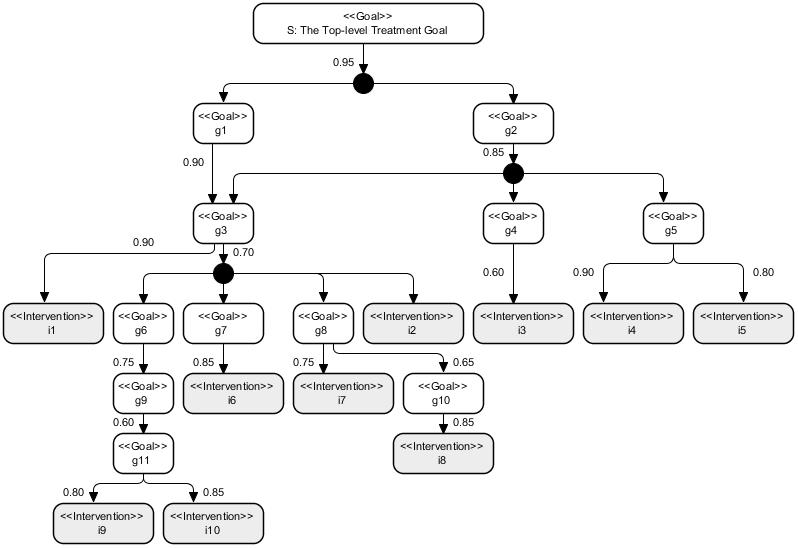}
    \caption{a Sample Treatment Model}
    \label{fig_tm}
\end{figure*}

The process of treatment modeling in \textit{GOTM} starts with identification of high-level treatment goals from existing treatment plans. Then high-level treatment goals are extracted and refined into lower-level treatment goals and/or medical interventions. The refinement process continues until all of the treatment goals are refined to achievable medical interventions~\cite{mougouei2012goal,mougouei2012measuring}. The treatment model of a sample medical guideline is depicted in Figure \ref{fig_tm}. Junction points in Figure \ref{fig_tm} represent logical AND-operator and their absence denotes OR-operator among treatment goals/medical interventions. As one example, the OR-operator between $i_9$ and $i_{10}$ in treatment model of Figure~\ref{fig_tm} indicates that in order to satisfy the treatment goal $g_{11}$, either of medical interventions $i_9$ or $i_{10}$ need to be employed. Also the AND-relation between treatment goals $g_1$ and $g_2$ specifies that both $g_1$ and $g_2$ are required for satisfaction of treatment goal $S$.

\subsection{Description of Treatment Guidelines}
\label{sec_gotm_description}

As discussed earlier, associations between treatment goals/ medical interventions are uncertain~\cite{pan_model_2015,ying_fuzzy_2006, belal_intelligent_2005,germino_uncertainty_1998,politi_communicating_2007}. In other words, treatment goals/medical interventions may partially contribute~\cite{mougouei2012measuring,mougouei2012evaluating,mougouei2015partial} to satisfaction of their higher-level treatment goals in a treatment model. In order to care for this partiality, we have employed a fuzzy-based description process~\cite{mougouei2013fuzzy1,mougouei2014fuzzy} for describing associations between treatment goals/medical interventions in a treatment model. The description process employed was first introduced by Mougouei et. al in ~\cite{mougouei2013fuzzyBased} for capturing uncertainty in modeling and description of security requirements~\cite{mougouei2012goal}. The process was later successfully adopted for capturing partiality in visibility requirement engineering of commercial websites~\cite{mougouei2014visibility}. The main components of our description process are explained in the following sections. 

\subsubsection{Construction of Fuzzy Grammar of Treatment}
\label{sec_gotm_description_construction}

A fuzzy grammar~\cite{mordeson2002fuzzy} of treatment will be constructed to describe treatment model of a medical disorder. Such grammar is identified by a quintuple of $GR = (s ,G , I, P, \mu)$ where $s$ denotes the top-level treatment goal, $G$ is the set of treatment goals, $I$ is the set of treatment requirements and $P$ specifies the set of \textit{Fuzzy Derivation Rules (FDRs)}. 

For treatment model of Figure~\ref{fig_tm}, $G =\{s, g_1,..., g_{11}\}$, $I = \{i_1,...,i_{10}\}, P = \{p_1,…,p_{16}\}$ and $S$ = ``top-level treatment goal''. The membership function  $\mu$ specifies the membership value of \textit{FDRs} as given by (\ref{eq_membership}). $d$ in (\ref{eq_membership}) denotes the degree to which a treatment goal (requirement) $r$ contributes to satisfaction of a treatment goal $g$.

\begin{eqnarray}
\label{eq_membership}
\mu (g \rightarrow r)= \mu (r,g) = d \in [0,1] 
\end{eqnarray}

\textit{FDRs} can be identified based on relations among treatment goals/requirements in treatment models of medical disorders. For instance, the derivation rule $p_1(s \rightarrow g_{1}g_{2})$ in Table~\ref{table_data} explains that $g_1$ and $g_2$ in treatment model of Figure~\ref{fig_tm} contribute to satisfaction of top-level treatment goal $s$. In addition, the fuzzy membership value of $p_1 (0.95)$ specifies that ``$g_1$ AND $g_2$'' can almost completely satisfy the treatment goal $s$. The membership values of \textit{FDRs} will be specified by their corresponding fuzzy membership functions. For instance, $\mu(s, g_{1}g_{2}) = \mu(s, g_1) = \mu(s, g_2) = 0.95$. 

\begin{table}[!h]
\caption{Derivation Rules for TM of Figure~\ref{fig_tm}}
\label{table_data}
\centering
\Large
\rowcolors{1}{}{lightgray}
\resizebox {0.5\textwidth }{!}{\begin{tabular}{  l  l  l  l  }
\Xhline{4\arrayrulewidth}
	\textbf{Rule}  & \textbf{Membership Value} & \textbf{Rule} & \textbf{Membership Value} \\ \hline 
	 
	$p_1: s \rightarrow g_{1}g_{2}$ & $0.95$ & $p_9: g_6 \rightarrow  g_9$ & $0.75$ \\ 
	$p_2: g_1 \rightarrow g_3$ & $0.9$ & $p_{10}: g_7 \rightarrow  i_6$ & $0.85$ \\ 
	$p_3: g_2 \rightarrow g_{3}g_{4}g_{5}$ & $0.85$ & $p_{11}: g_8 \rightarrow  i_7$ & $0.75$ \\ 
	$p_4: g_{3}\rightarrow i_1$ & $0.9$ & $p_{12}: g_8 \rightarrow  g_{10}$ & $0.65$ \\ 
	$p_5: g_3 \rightarrow g_{6}g_{7}g_{8}i_{2}$ & $0.7$ & $p_{13}: g_9 \rightarrow  g_{11}$ & $0.6$ \\ 
	$p_6: g_{4} \rightarrow i_{3}$ & $0.6$ & $p_{14}: g_{10} \rightarrow  i_{8}$ & $0.85$ \\ 
	$p_7: g_{5}\rightarrow  i_{4}$ & $0.9$ & $p_{15}: g_{11} \rightarrow  i_{9}$ & $0.8$ \\ 
	$p_8: g_{5} \rightarrow i_{5}$ & $0.8$ & $p_{16}: g_{11} \rightarrow  i_{10}$ & $0.85$ \\ 
	\Xhline{4\arrayrulewidth}
\end{tabular}
}%

\end{table}

\subsubsection{Calculation of Impact}
\label{sec_gotm_description_impact}

Let $r_1,...,r_n$ to be members of $(G \cup R)^*$ , then $r_1 \rightarrow ... \rightarrow r_n$ specifies a derivation chain for $r_n$. For a pair of $(v,g)$ in which $v$ denotes a treatment requirement and $g$ denotes a treatment goal on a derivation chain of $v$, impact is given by
the membership function $\mu^\infty(g,v)$ where $\mu^\infty(g,v)$ specifies the strongest contribution of $v$ to satisfaction of goal $g$. 

The value of $\mu^\infty(g,v)$ is calculated through taking supremum over membership values of all derivation chains which generate $v$. The membership value of a derivation chain is calculated by taking minimum over membership values of all
FDRs on that derivation chain. Equation~(\ref{eq_impact}) demonstrates the calculation of impact. Fuzzy operators $\vee$ and $\wedge$  denote fuzzy OR (maximum) and fuzzy AND (minimum) operators respectively~\cite{mougouei2013fuzzy}. 

\vspace{1em}
\begin{align}
\label{eq_impact}
&\forall p_i=(n_0,...,n_k) \in P, \mu(p_i) = \bigwedge_{j=1}^{k}\text{ }\mu(n_{j-1},n_j)\\ \nonumber
&\mu^{\infty}(n_0,n_k) = \bigvee_{i=1}^{m}\mu(p_i)
\vspace{1em}
\end{align}

As one example for calculation of impact, $\mu(i_5)$ is calculated for the requirement $i_5$ in treatment model of Figure~\ref{fig_tm}. For $i_5$ to be generated, there are two derivation chains in treatment model of Figure~\ref{fig_tm}: $1)s \rightarrow g_1 \rightarrow g_3 \rightarrow i_1$, and $2) s \rightarrow g_2 \rightarrow g_3 \rightarrow r_1$. And so, we  have: $\mu^\infty(s,i_1)=(0.95\wedge 0.9 \wedge 0.9)\vee(0.95 \wedge 0.85 \wedge 0.9)$. 

The impact values for treatment requirements of Figure~\ref{fig_tm} are listed in Table~\ref{table_impact}. For each treatment requirement $r$, different impact values are calculated with respect to different treatment goals on the derivation chains of $r$. For instance, $\mu^\infty(s,i_6) = 0.7$, and $\mu^\infty(g_7,i_6) = 0.85$. 

\begin{table}[!h]
\caption{Calculated Impact Values for Treatment Requirements of Figure~\ref{fig_tm}}
\label{table_impact}
\centering
\Large
\rowcolors{1}{}{lightgray}
\resizebox {0.5\textwidth }{!}{
\begin{tabular}{  l  l l  l  l  l  l  l  l  l  l  }
\toprule[1.5pt]
	\textbf{Goal} & $\mu^\infty(i_1)$ & $\mu^\infty(i_2$) & $\mu^\infty(i_3)$ & $\mu^\infty(i_4)$ & $\mu^\infty(i_5)$ & $\mu^\infty(i_6)$ & $\mu^\infty(i_7)$ & $\mu^\infty(i_8)$ & $\mu^\infty(i_9)$ & $\mu^\infty(i_{10})$ \\ \hline 
	$s$ & 0.9 & 0.7 & 0.6 & 0.85 & 0.8 & 0.7 & 0.7 & 0.65 & 0.6 & 0.6 \\ 
	$g_1$ & 0.9 & 0.7 & 0 & 0 & 0 & 0.7 & 0.7 & 0.65 & 0.6 & 0.6 \\ 
	$g_2$ & 0.85 & 0.7 & 0.6 & 0.85 & 0.8 & 0.7 & 0.7 & 0.65 & 0.6 & 0.6 \\ 
	$g_3$ & 0.9 & 0.7 & 0 & 0 & 0 & 0.7 & 0.7 & 0.65 & 0.6 & 0.6 \\ 
	$g_4$ & 0 & 0 & 0.6 & 0 & 0 & 0 & 0 & 0 & 0 & 0 \\ 
	$g_5$ & 0 & 0 & 0 & 0.9 & 0.8 & 0 & 0 & 0 & 0 & 0 \\ 
	$g_6$ & 0 & 0 & 0 & 0 & 0 & 0 & 0 & 0 & 0.6 & 0.6 \\ 
	$g_7$ & 0 & 0 & 0 & 0 & 0 & 0.85 & 0 & 0 & 0 & 0 \\ 
	$g_8$ & 0 & 0 & 0 & 0 & 0 & 0 & 0.75 & 0.65 & 0 & 0 \\ 
	$g_9$ & 0 & 0 & 0 & 0 & 0 & 0 & 0 & 0 & 0.6 & 0.6 \\ 
	$g_{10}$ & 0 & 0 & 0 & 0 & 0 & 0 & 0 & 0.85 & 0 & 0 \\ 
	$g_{11}$ & 0 & 0 & 0 & 0 & 0 & 0 & 0 & 0 & 0.8 & 0.85 \\ 
\bottomrule[1.5pt]
\end{tabular}
}%

\end{table}

The impacts can be used to develop an optimization model that finds an optimal subset of medical interventions with highest effectiveness, that takes into account the influences of those interventions on the effectiveness of one another~\cite{mougouei2017integer}. Also, mathematical techniques can be employed~\cite{moeini2017identification} to identify constraints that can enhance the execution time of such optimization model. When collecting large samples of impacts is not feasible, those impacts can be re-sampled based on the distribution of the collected samples~\cite{moeini2011conditional,ahmadi2016economical}.

\section{Conclusion and Future Work}
\label{sec_conclusion}

This paper proposed caring for uncertainty in treatment modeling of medical disorders. To this end, we have proposed using a fuzzy-based technique for modeling and description of treatment models. The algebraic structure of the proposed modeling and description technique allows for capturing uncertainties of associations between goals/requirements of treatment guidelines. We have further contributed a goal-oriented framework which allows for goal-oriented modeling and description of treatment guidelines. Goal-oriented approach toward modeling treatment guidelines in \textit{GOTM} provides structured arguments to choose from medical requirements those satisfying particular treatment goals. 

\bibliographystyle{IEEEtran}
\bibliography{ref}
\end{document}